\documentclass{article}
\usepackage[dblblindworkshop,final]{neurips_2026}
\workshoptitle{Interpreting Agent Behavior}
\IfFileExists{phvr7t.tfm}{}{\renewcommand{\sfdefault}{cmr}}

\usepackage{amsmath,amsfonts,bm}

\def\eqref#1{equation~\ref{#1}}

\def\1{\bm{1}}

\DeclareMathAlphabet{\mathsfit}{\encodingdefault}{\sfdefault}{m}{sl}
\SetMathAlphabet{\mathsfit}{bold}{\encodingdefault}{\sfdefault}{bx}{n}

\usepackage{url}
\usepackage{booktabs}
\usepackage{graphicx}
\usepackage{float}
\usepackage{afterpage}
\usepackage{amsmath,amssymb,amsthm}
\usepackage{microtype}
\usepackage{xcolor}
\usepackage{hyperref}
\hypersetup{hidelinks}
\IfFileExists{wrapfig.sty}{\usepackage{wrapfig}}{\IfFileExists{tex/wrapfig.sty}{\usepackage{tex/wrapfig}}{\newenvironment{wrapfigure}[3][]{\begin{figure}[t]\centering\begin{minipage}{##3}\centering}{\end{minipage}\end{figure}}}}

\title{RetailAgent: Structured Adverse Timing in Self-Conditioned Multimodal LLM Trading Agents}

\author{%
  Yupeng Zhang$^{1*}$,\quad
  Liuyuan Jiang$^{2*}$,\quad
  Hongyi Huang$^{1}$,\quad
  Bingheng Li$^{3}$,\quad
  Lisha Chen$^{2}$ \\
  $^{1}$University of Wisconsin-Madison \quad
  $^{2}$University of Rochester \quad
  $^{3}$Michigan State University \\
  \texttt{yupeng.zhang@wisc.edu} \quad
  \texttt{ljiang24@ur.rochester.edu} \quad
  \texttt{hhuang377@wisc.edu} \\
  \texttt{libinghe@msu.edu} \quad
  \texttt{lisha.chen@rochester.edu}
}

\begin{document}

\maketitle
\begingroup
\renewcommand{\thefootnote}{\fnsymbol{footnote}}
\footnotetext[1]{Equal contribution.}
\endgroup

\begin{abstract}
In financial markets, a sequential policy that reacts systematically to price movements may become predictable to other market participants. This paper studies whether large language model (LLM) agents exhibit such directional structure through \emph{RetailAgent}, an experimental framework in which an LLM observes anonymized intraday equity price histories and permitted state, then repeatedly chooses \emph{long} (hold the stock) or \emph{flat} (stay out) before the subsequent interval return is revealed. We compare returns during long and flat intervals along the same stock's intraday path after removing the overall fraction of long decisions. This exposure-matched measure reveals persistent negative timing across modality, horizon, state, and model family. Shuffling saved action sequences substantially attenuates the effect, showing that alignment between actions and subsequent returns drives the negative score. Feeding self-authored memories into decisions further increases policy persistence, while timing becomes more negative among stock-days on which the agent uses both actions. These results reveal stable, recoverable directional structure in sequential LLM financial decisions and a behavioral signal for studying how another participant could respond to a predictable policy.
\end{abstract}

\section{Introduction}
\label{sec:intro}

\begin{wrapfigure}{r}{0.4\textwidth}
 \centering
 \vspace{-10mm}
 \includegraphics[width=0.93\linewidth,height=5.3cm]{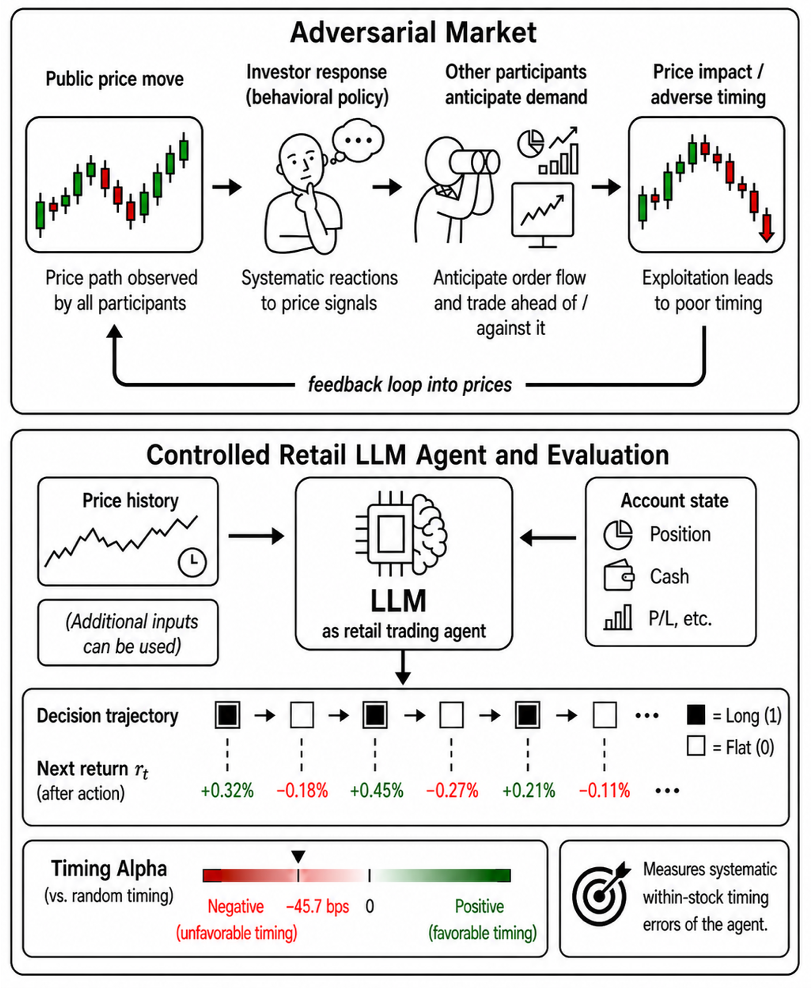}
 \vspace{-3mm}
\caption{Top: price-conditioned behavior may invite a counter-response in an interactive market. Bottom: RetailAgent isolates a policy, chooses a binary action from permitted state, and is evaluated on revealed returns.}
 \label{fig:intro-dynamics}
 \vspace{-5.2mm}
\end{wrapfigure}

Financial-market theory shows that prices can aggregate dispersed information and that order flow, the sequence of submitted trades, can reveal information to other market participants \citep{grossman1980impossibility,kyle1985continuous}. A policy that responds systematically to salient price movements can therefore be anticipated even when each action sounds plausible in isolation. Evidence connects individual trading to attention, past returns, market displays, feedback, overconfidence, and costly turnover~\citep{barber2000trading,barber2008attention,barber2013behavior,benamar2019see,statman2006investor,ben2018uninformative}. Stocks favored by small traders have also displayed subsequent underperformance in historical samples \citep{hvidkjaer2008small}. The top panel of Figure~\ref{fig:intro-dynamics} illustrates how systematic price-conditioned reactions create behavior that other market participants can anticipate.

Large language model (LLM) agents offer a controlled setting for studying such behavior because they make sequential decisions conditioned on prior actions, stored memories, and self-generated interpretations \citep{park2023generative,liu2024agentbench}. The bottom panel of Figure~\ref{fig:intro-dynamics} isolates one such policy. An aggregate task score can combine irregular mistakes with a repeated tendency to choose the wrong side of subsequent returns. We isolate that repeated tendency under a \emph{fixed information boundary}. The agent receives permitted price history and endogenous state generated by earlier decisions. The principal numerical protocol conceals stock identity, calendar date, absolute price, news, fundamentals, cross-sectional ranks, and future outcomes. The appendix records the separate input provenance of visual and learned conditions. This design evaluates the long/flat policy on a controlled market path \citep{aher2023using,argyle2023outofone,hu2026simbench}.

Building on this controlled setup, we make the following contributions, summarized as C1--C3.

\noindent\textbf{C1) RetailAgent framework.}
We propose \emph{RetailAgent}, a controlled multimodal framework for tracing sequential LLM decisions and self-authored memory under that boundary (Section~\ref{sec:method}).

As illustrated in the bottom panel of Figure~\ref{fig:intro-dynamics} and detailed in Figure~\ref{fig:architecture}, the agent receives an anonymized rolling intraday price history and permitted endogenous state through multimodal pathways, then chooses \emph{long} or \emph{flat}. At each interval, the framework records the observation, endogenous state, action, any self-authored memory, and subsequently revealed return. RetailAgent therefore provides a controlled behavioral proxy with a limited price-only interface. We ask whether sequential LLM decisions exhibit stable directional structure. The experiments identify a persistent wrong-signed component:

\noindent\textbf{C2) Verified structured adverse timing.}
We establish consistently negative within-stock timing across 14 standard configurations and use trajectory shuffling, the complementary schedule, and signal-overlap tests to make this directional structure explicit (Section~\ref{sec:experiments}).

Across all 14 standard configurations spanning modality, decision horizon, account state, and model family under a shared protocol, within-stock timing is negative, indicating systematically wrong-signed entry and exit relative to each stock's own path (Section~\ref{sec:standard-timing}). In the principal 10-minute {Qwen3.5} experiments~\citep{qwen35_9b}, numerical-text, chart-only, and joint text+chart inputs score $-45.7$~bps per stock-day on 13,710 stock-days, $-29.9$ on 14,937, and $-48.9$ on 14,438. These panels use different sampling frames. The repeated negative sign is the robust claim, and magnitude comparisons use matched panels.

Trajectory shuffling substantially attenuates the effect. Randomly permuting actions while preserving the overall long fraction yields $-3.5$~bps under global shuffling and $-8.7$ under same-day shuffling, compared with $-45.7$ for the intact schedule. Sequential alignment between actions and subsequent returns therefore accounts for most of the intact estimate. The complementary schedule, formed by exchanging long and flat actions, reverses the timing sign by construction and serves as a sign diagnostic. Under research-return labels, the trajectory carries recoverable directional structure. Section~\ref{sec:overlap} shows that projections onto the tested GRU, GBDT, and reversal signals leave substantial residual structure in the inverted schedule.

\afterpage{%
\begin{figure*}[t]
 \centering
 \includegraphics[width=0.9\textwidth,height = 6.3cm]{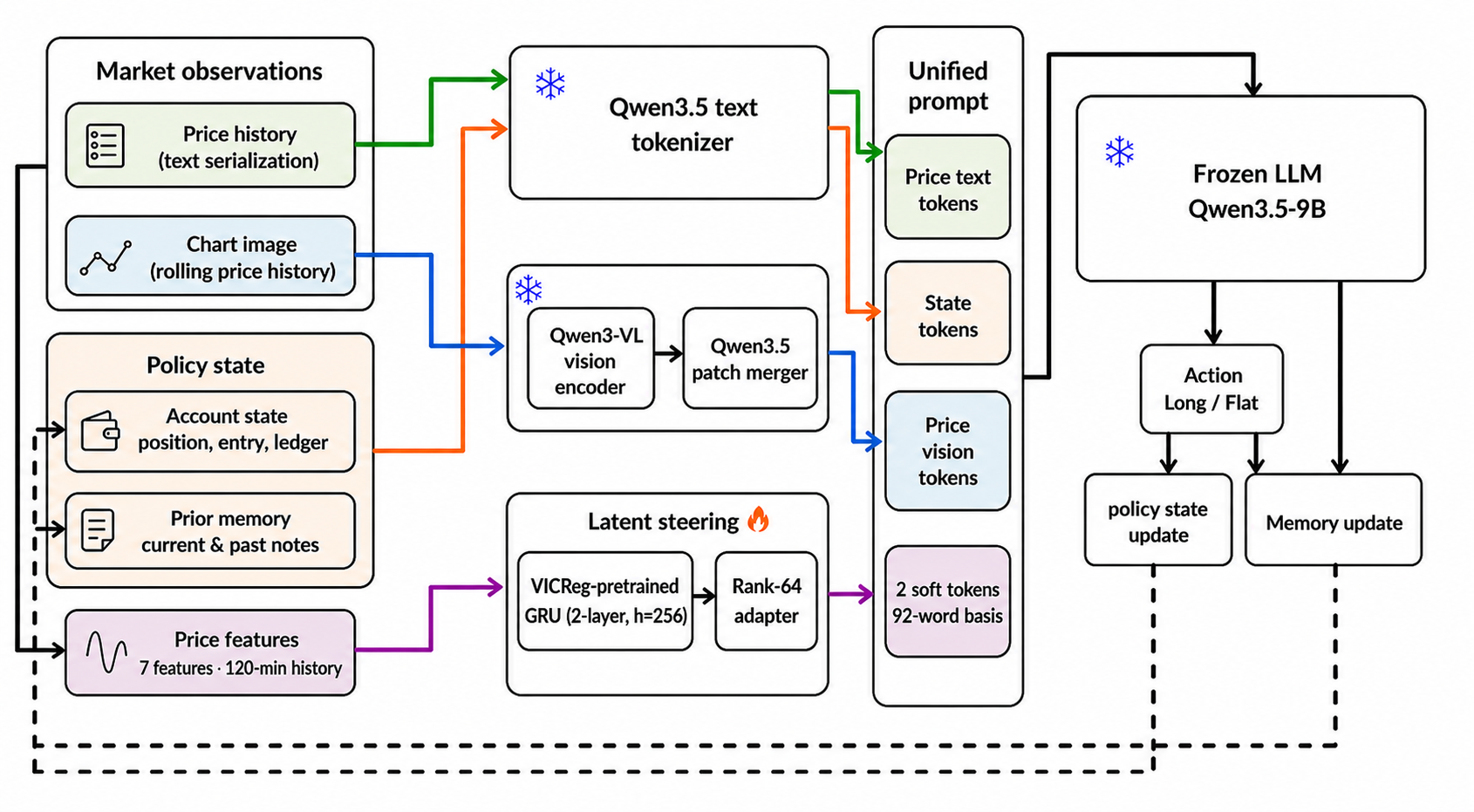}
 \vspace{-0.4cm}
  \caption{\textbf{RetailAgent architecture.} The locally cached \texttt{Qwen/Qwen3.5-9B} processor handles price text, account state, and charts through its text and vision pathways~\citep{qwen35_9b,bai2025qwen3vl}. The dashed route carries self-authored memory into later decisions. The learned branch maps seven price features through a VICReg-pretrained GRU encoder~\citep{bardes2021vicreg,cho2014learning} and low-rank adapter into two continuous embeddings~\citep{li2021prefix,lester2021power} on a 92-word basis. The frozen LLM chooses long or flat before the return is revealed.}
  \vspace{-0.3cm}
 \label{fig:architecture}
\end{figure*}}

\noindent\textbf{C3) Self-conditioning amplifies persistence and adverse timing.}
We show that exposing one earlier self-authored memory reduces action switching and makes within-stock timing more negative among stock-days that contain both actions (Section~\ref{sec:self-conditioning}).

Along the dashed route in Figure~\ref{fig:architecture}, the agent writes a memory for its future self and includes that language in the next decision context. The memory window $w$ counts how many earlier self-authored memories are visible at decision time. Among scored stock-days with both actions, moving from $w=0$ (current memory only) to $w=1$ (one prior memory also visible) shifts timing from $-62.8$ to $-74.1$~bps and position changes from $2.5$ to $1.7$ per stock-day. Across all attempted stock-days, the corresponding action traces also show greater persistence under $w=1$. Continuous embedding state provides a complementary comparison by conditioning the frozen model through price-derived soft tokens. Section~\ref{sec:counter-policy} characterizes the resulting directional structure through the complementary schedule and the timing left after removing fitted linear price-signal components.

\subsection{Related work}
\label{sec:related}

Financial LLM research spans domain models, benchmarks, and methods for business reasoning, financial mathematics, market-movement prediction, expert search, and verbal price-series analysis~\citep{wu2023bloomberggpt,liu2023fingpt,krumdick2024bizbench,xie2024finben,zhao2024financemath,saqur2025filtered,hu2026finsearchcomp,koa2026reasoning}. Trading agents add memory, reflection, and specialized roles across sequential decisions~\citep{yu2025finmem,yu2024fincon,xiao2024tradingagents}. Related work also studies LLMs as experimental agents and conditions frozen models with learned prompts~\citep{park2023generative,liu2024agentbench,aher2023using,argyle2023outofone,hu2026simbench,li2021prefix,lester2021power}. RetailAgent complements these lines by auditing a fixed-path long/flat action trace, varying visible self-authored memory within the same policy, and comparing linguistic conditioning with price-derived soft tokens. Exposure-matched within-stock timing and persistence reveal predictable directional structure beyond aggregate task and portfolio outcomes.

\section{Methodology}
\label{sec:method}

This section defines the timing metrics and how RetailAgent constructs sequential decisions under a fixed information boundary. Section~\ref{sec:background} introduces the preliminary knowledge and settings, and Sections~\ref{sec:controlled-decisions}--\ref{sec:narrative-latent} describe the decision loop and conditioning channels.

\subsection{Setting and metrics}
\label{sec:background}

RetailAgent operates on anonymized equity price paths over an intraday session. The experimental unit is a stock-day pair $(i,d)$, where $i$ indexes the anonymized stock and $d$ the trading date. Each scored intraday grid contains $T$ sequential decision intervals, indexed by $t\in\{1,\ldots,T\}$. At the start of interval $t$, the agent observes condition-specific price history and chooses a binary action $p_{idt}\in\{0,1\}$, where $1$ denotes \emph{long} (hold over the next interval) and $0$ denotes \emph{flat} (stay out). The immediately following interval has research-return label $r_{idt}$, which is the relative price change over that interval. The principal numerical protocol reveals this label after the action and conceals stock identity, calendar date, absolute price level, news, fundamentals, and cross-sectional ranks. One basis point (bp) equals $0.01\%$. \emph{Exposure} $\bar p_{id}$ is the fraction of intervals on the stock-day in which the agent is long.

For a stock-day with $T$ decision intervals and return vector $r_{id}=(r_{id1},\ldots,r_{idT})$, \emph{timing alpha} is the exposure-matched within-stock timing
\begin{equation}
 A_{id}(p,r)=\sum_t\left(p_{idt}-\bar p_{id}\right)r_{idt},
 \qquad \bar p_{id}=T^{-1}\sum_t p_{idt}.
 \label{eq:alpha}
\end{equation}
After removing average exposure, $A_{id}$ measures whether the agent is long during relatively favorable intervals on the \emph{same} stock-day. The same definition implies the decomposition
\begin{equation}
 \sum_t p_{idt}\, r_{idt}=\bar p_{id}\sum_t r_{idt}+A_{id}(p,r),
 \label{eq:decomp}
\end{equation}
so long-only cumulative return on the stock-day equals passive exposure to the return path plus the centered timing term. Thus, positive task-level return can coexist with systematically adverse timing. \emph{Primary metric.} We report the mean of $A_{id}$ in bps per stock-day. \emph{Sign.} Positive values indicate directionally aligned timing, values near zero indicate noise-like timing, and negative values indicate wrong-signed timing. A policy that is always long scores zero by construction. \emph{Sign diagnostic.} The complementary schedule $q_{idt}=1-p_{idt}$ satisfies $A_{id}(q,r)=-A_{id}(p,r)$ by definition.

\noindent\textbf{Example.} On a saved trajectory with one earlier self-authored memory visible ($w=1$), long-interval return of $+359.4$~bps coexists with timing alpha of $-867.3$~bps, while flat intervals return $+1{,}404.0$~bps. The agent remained flat through much of a strong rise and entered before comparatively weaker intervals. This trace shows how a positive long-only outcome can accompany adverse within-stock timing.

The cross-sectional information coefficient (IC) measures contemporaneous stock selection by correlating positions and subsequent returns across the set $S_{dt}$ of stocks available to a condition at the same decision time,
\begin{equation}
 \mathrm{IC}_{dt}=\operatorname{Corr}_{i\in S_{dt}}(p_{idt},r_{idt}). \label{eq:IC}
\end{equation}
Timing measures \emph{when} a stock is held, while \eqref{eq:IC} measures \emph{which} stocks are preferred. Brokerage evidence on retail attention, overconfidence, turnover, and cross-sectional return patterns~\citep{barber2000trading,barber2008attention,statman2006investor,ben2018uninformative,benamar2019see,hvidkjaer2008small} motivates evaluating both margins under a sparse price-and-state interface. RetailAgent provides a controlled policy proxy, with retail-population behavior as a separate empirical target.

\subsection{Controlled decisions and evaluation}
\label{sec:controlled-decisions}

RetailAgent separates the observation supplied to an agent from the endogenous state through which it interprets that observation. Figure~\ref{fig:architecture} summarizes the text and vision pathways, the dashed self-conditioning route, recursive policy state, and optional latent-steering branch. Across conditions we hold fixed the binary long/flat action semantics and research-return scoring. We vary observation construction and the endogenous state carried forward.

At the start of interval $t$, the agent receives condition-specific observation $x_{id,1:t}$, presented as numerical text, a chart, or both, together with endogenous state $z_{idt}^{(c)}$ such as account variables, memory history, or soft tokens. These inputs enter the unified prompt that feeds the frozen LLM (cf.\ Figure~\ref{fig:architecture}). The agent then emits a binary position $p_{idt}\in\{0,1\}$, where $1$ denotes long and $0$ denotes flat. The protocol reveals the research-return label $r_{idt}$ after the action. The next interval repeats with updated observation and endogenous state. Over the day this yields a trajectory of actions, optional self-authored memories, and subsequently revealed labels. Matched comparisons use the same observation path and vary the specified carried state.

We score trajectories with timing $A_{id}(p,r)$ and contemporaneous IC (\eqref{eq:IC}). Timing measures entry and exit quality on a stock's path. IC measures cross-sectional ranking at the same decision time. The complementary schedule $1-p$ flips the timing sign by construction. Each condition induces $p_{idt}\sim\pi_c(\cdot\mid x_{id,1:t},z_{idt}^{(c)})$.

\subsection{Interventions and self-conditioning}
\label{sec:narrative-latent}

\emph{Conditioning} changes representation or endogenous state while keeping the market information set and binary action space fixed. We apply three families.

\textbf{Observation representation.} Numerical text serializes completed intervals, making local differences and ordering directly available to linguistic reasoning. The chart provides a fine-resolution visual path, making slope, curvature, extrema, and visual grouping jointly salient. Multimodal presents both views through the unified prompt. These conditions hold the underlying stock-day fixed while changing resolution, lookback, and representation.

\textbf{Consequential account state.} Stateful prompts progressively reveal current position and entry, a trade ledger, or an account summary. Position and entry supply immediate action state. The ledger records prior commitments, and the account summary presents accumulated action-weighted return and entry-relative performance. Because these variables depend on earlier actions, each decision changes part of the state that conditions the next decision and induces a recursive policy.

\textbf{Narrative and latent endogenous state.} Narrative self-conditioning uses a two-pass protocol on the dashed route in Figure~\ref{fig:architecture}. A first pass writes a short memory to the model's future self. A second pass conditions on the current observation, that memory, and up to $w$ earlier memories before choosing long or flat. We evaluate $w\in\{0,1,5,12\}$. At $w=0$, the decision uses the current-step memory and zero earlier memories. At $w\geq 1$, earlier commitments are visible. Directives are absent, emphasize conviction, or ask for consistency unless evidence has clearly changed. All narrative conditions use inference only.

An optional latent channel supplies continuous, price-dependent state:
\begin{equation}
 x_{id,1:t}\xrightarrow{f_\theta}h_{idt}
 \xrightarrow{g_\phi}(e_{1},\ldots,e_k)
 \longrightarrow \text{frozen LLM}\longrightarrow p_{idt}.
\end{equation}
Here $f_\theta$ is a gated recurrent unit (GRU) price encoder~\citep{cho2014learning}, $h_{idt}$ is its hidden state, $g_\phi$ is a low-rank transformation with rank 64, and $e_1,\ldots,e_k$ are soft-token embeddings~\citep{li2021prefix,lester2021power}. The GRU is pretrained with variance-invariance-covariance regularization (VICReg), a self-supervised objective~\citep{bardes2021vicreg}. It maps seven price features into $k{=}2$ embeddings, each formed as a weighted combination of 92 fixed market-related word embeddings. Ablations vary token count, basis size, encoder freezing, and free embedding vectors. Comparing narrative and latent conditioning tests how linguistic self-commitment and continuous state affect persistence.

\section{Experiments}
\label{sec:experiments}

We evaluate RetailAgent across modality, horizon, account state, model family, and self-conditioning interventions. Section~\ref{sec:results} reports timing, shuffle and overlap controls, and narrative effects. Section~\ref{sec:counter-policy} explains how the recovered directional structure could inform another participant's response and states the empirical scope.

\subsection{Experimental setup}

Appendix~\ref{app:protocol} provides detailed documentation of panel construction, prompts, generation and parsing rules, optimization, statistical inference, saved artifacts, and computational resources.

\textbf{Market panels and conditions.} CSI-500 is a mid-cap Chinese equity index~\citep{csi500}. The source data provide each anonymized stock-day on a 239-minute intraday grid. Numerical conditions receive indexed histories constructed from completed non-overlapping intervals. Fine charts and encoder features use the archived one-minute reconstruction. The primary panel samples 23 decisions at 10-minute intervals, with 5- and 20-minute panels as horizon checks. Outcomes are subsequent non-overlapping research-return labels. Narrative and encoder evaluations attempt 1,500 stock-days, while larger trajectory sets support modality, state, and horizon checks. The principal model is {Qwen3.5-9B}~\citep{qwen35_9b}. Sign checks use {Claude Haiku} 4.5~\citep{anthropic2025haiku} and {Granite-4.0-H-Small}~\citep{granite2025}. Stateful variants add position and entry, a trade ledger, or an account summary. Memory conditions use $w\in\{0,1,5,12\}$ and absent, conviction, or consistency directives. Standard and narrative conditions use inference only. Learned encoder and adapter conditions include an optimization stage.

\textbf{Generation and scoring.} Principal Qwen runs use vLLM~\citep{kwon2023efficient} in bfloat16 with thinking disabled, temperature 0.7, and top-$p$ 0.95. Text and state outputs use 160 tokens, chart and joint outputs use 320, and hosted outputs use 350. Memory uses 220 tokens followed by a greedy 24-token decision. Local standard, narrative, and hosted generation paths apply their recorded parsers and failure rules. Timing estimates retain trajectories containing both actions, and confidence intervals resample dates. The optional encoder is a two-layer GRU with 256 units, a 120-minute lookback, seven archived channels, and a rank-64 adapter into two soft tokens. Phase-two runs use a fixed initial decision bias and an exposure-collapse penalty. The memory-conditioned phase-three run targets long exposure 0.40 through an online bias update. Learned conditions use hard actions with soft backward gradients. Training dates occupy the first 3/5 of the applicable panel, where outcome transformations are estimated. The saved memory-conditioned checkpoint is evaluated on 1,500 stock-days from 28 later dates.

\textbf{Statistical inference and controls.} Code, trajectories, filenames, and panels preserve the prompts, actions, selected memories, failure counters, model identifiers, and labels. Timing aggregates within stock-day and then averages in bps. The primary null draws exposure-matched random schedules, giving near-zero timing by construction. Percentile bootstraps resample trading dates. Cross-sectional IC uses decision times with position variation across stocks. For each comparator $s$, overlap tests remove its within-stock linear component from $q=1-p$ and score the remaining timing.
\subsection{Results}
\label{sec:results}

Across every standard configuration, RetailAgent enters and exits at comparatively unfavorable intervals. Matched-panel analyses then examine the pattern's sequential structure and response to self-conditioning.

\subsubsection{Standard timing and controls}
\label{sec:standard-timing}

Table~\ref{tab:main-results} consolidates 14 negative timing estimates across three decision horizons, four state conditions, text and visual inputs, and two model families. The principal 10-minute conditions score $-45.7$~bps for text, $-29.9$ for chart, and $-48.9$ for multimodal inputs. Their distinct sampling frames support a descriptive sign comparison.
The shuffle controls isolate trajectory alignment on the principal text condition. Global shuffling yields $-3.5$~bps and same-day shuffling yields $-8.7$~bps, compared with $-45.7$ for the intact schedule. These controls preserve the overall long fraction while breaking sequential alignment. The intact trajectory therefore contains substantially more negative timing. Haiku and Granite provide cross-model sign checks. Granite retains the negative sign across four state prompts on 600 stock-days per prompt. Appendix~\ref{app:additional} provides additional visual estimates and detailed model records.

\begin{table*}[t]
 \centering
 \footnotesize
 \setlength{\tabcolsep}{4.5pt}
 \renewcommand{\arraystretch}{0.96}
 \begin{tabular}{lllrrr}
 \toprule
 Model or control & Observation or state & Interval & Stock-days & Timing alpha, bps (95\% CI) & Switches \\
 \midrule
 \multicolumn{6}{l}{\textit{A. Fourteen standard configurations}} \\
 Qwen3.5 & price text & 10m & 13,710 & $-45.7\,[-49.4,-41.8]$ & \\
 Qwen3.5 & chart & 10m & 14,937 & $-29.9\,[-33.7,-26.2]$ & \\
 Qwen3.5 & multimodal & 10m & 14,438 & $-48.9\,[-52.7,-45.0]$ & \\
 Qwen3.5 & price text & 20m & 23,594 & $-56.7\,[-59.8,-53.6]$ & \\
 Qwen3.5 & +position & 20m & 25,410 & $-54.7\,[-58.2,-51.0]$ & \\
 Qwen3.5 & +ledger & 20m & 25,718 & $-43.9\,[-47.0,-40.8]$ & \\
 Qwen3.5 & +account & 20m & 22,013 & $-51.1\,[-55.0,-46.7]$ & \\
 Qwen3.5 & price text & 5m & 9,937 & $-38.8\,[-42.9,-34.6]$ & \\
 Qwen3.5 & +position & 5m & 9,961 & $-44.8\,[-49.3,-40.4]$ & \\
 Qwen3.5 & +ledger & 5m & 9,966 & $-30.5\,[-33.5,-27.2]$ & \\
 Qwen3.5 & +account & 5m & 9,052 & $-53.2\,[-56.2,-50.1]$ & \\
 Claude Haiku & price text & 5m & 827 & $-38.8\,[-52.9,-22.5]$ & \\
 Claude Haiku & price text & 10m & 165 & $-42.9\,[-62.0,-24.8]$ & \\
 Qwen3-VL & coarse image & 20m & 28,387 & $-32.7\,[-35.7,-29.8]$ & \\
 \midrule
 \multicolumn{6}{l}{\textit{B. Trajectory and self-conditioning controls}} \\
 Global shuffle & price-text actions & 10m & 13,716 & $-3.5\,[-5.4,-1.7]$ & \\
 Same-day shuffle & price-text actions & 10m & 13,716 & $-8.7\,[-10.4,-7.1]$ & \\
 Qwen3.5, $w=0$ & current memory & 10m & 1,194 & $-62.8\,[-69.7,-56.0]$ & $2.5$ \\
 Qwen3.5, $w=1$ & +1 earlier memory & 10m & 1,054 & $-74.1\,[-81.5,-66.6]$ & $1.7$ \\
 \midrule
 \multicolumn{6}{l}{\textit{C. Granite cross-model sign checks}} \\
 Granite & price text & 10m & 600 & $-36.3$ & \\
 Granite & +position & 10m & 600 & $-25.6$ & \\
 Granite & +ledger & 10m & 600 & $-14.1$ & \\
 Granite & +account & 10m & 600 & $-8.2$ & \\
 \bottomrule
 \end{tabular}
 \caption{Exposure-matched timing alpha. Principal local text, chart, and multimodal rows use \texttt{Qwen/Qwen3.5-9B}~\citep{qwen35_9b}. The coarse-image archive is recorded as Qwen3-VL~\citep{bai2025qwen3vl}. Other models are Claude Haiku 4.5~\citep{anthropic2025haiku} and Granite-4.0-H-Small~\citep{granite2025}. Confidence intervals resample dates. Switches count long/flat changes per stock-day. Granite rows are archived point estimates. Cross-frame magnitudes are descriptive.}
 \label{tab:main-results}
\end{table*}

The 20-minute state rows range from $-43.9$ to $-56.7$~bps, and the 5-minute rows range from $-30.5$ to $-53.2$~bps. Their changing order shows that additional account information changes several state inputs at once. Every state and horizon configuration retains the negative sign.
The Haiku result reproduce that sign on smaller samples, including 165 stock-days at 10 minutes. Their role is a model-family sign check with lower statistical power. Granite extends the check across price text, position, ledger, and account prompts. Each Granite row uses 600 stock-days, and its point estimates range from $-8.2$ to $-36.3$~bps. These cross-model rows support sign recurrence. The sampling differences make their magnitude comparisons descriptive.

On the principal text sample, the difference between the LLM schedule and exposure-matched random schedules is $-46.2$~bps with 95\% CI $[-50.5,-41.8]$. Price-level and last-return schedules have their own timing patterns. We therefore use random exposure-matched schedules as the behavioral benchmark and price-based signals as overlap comparators.

\subsubsection{Signal overlap with price baselines}
\label{sec:overlap}

The complementary schedule makes the sign of the behavior explicit and is algebraically determined by \eqref{eq:alpha}. We measure its linear overlap with three conventional price signals. The GRU comparator is a gated recurrent unit price model. GBDT-38 is a gradient-boosted decision tree using 38 price features. One-period reversal takes the opposite side of the previous interval's return. For each comparator, we fit its within-stock linear relation to the inverted position and score the unexplained remainder. Table~\ref{tab:projection} shows that the remainder retains substantial positive timing.

\begin{table}[t]
 \centering
 \small
 \begin{tabular}{lrrrr}
 \toprule
 Comparator & Stock-days & Corr. with $q$ & Inverted & Residual timing (95\% CI) \\
 \midrule
 GRU price model & 2,519 & $+0.114$ & $+46.8$ & $+43.9\,[38.1,49.5]$ \\
 GBDT-38 & 13,717 & $+0.132$ & $+45.7$ & $+40.4\,[37.0,43.8]$ \\
 One-period reversal & 13,717 & $+0.286$ & $+45.7$ & $+38.9\,[36.0,41.8]$ \\
 \bottomrule
 \end{tabular}
 \caption{Inverted text-agent timing after within-stock projection on conventional price signals.} 
 \label{tab:projection}
\end{table}

Correlations with the inverted schedule remain below $+0.3$, yet the GRU projection retains $43.9$ of $46.8$~bps, the GBDT projection retains $40.4$ of $45.7$~bps, and the reversal projection retains $38.9$ of $45.7$~bps. The date-clustered intervals are positive in all three rows. These residuals quantify the directional structure that remains after each linear projection under the research labels.

\subsubsection{Self-conditioning and persistence}
\label{sec:self-conditioning}

Narrative self-conditioning reorganizes temporal behavior under the same information boundary. Table~\ref{tab:memory-sweep} reports the complete ten-condition sweep and the later-period persistence evidence. Every conditional main-period timing interval lies below zero. Among stock-days with both actions, moving from current memory only ($w=0$) to one earlier memory ($w=1$) shifts timing from $-62.8$ to $-74.1$~bps and reduces position changes from $2.5$ to $1.7$ per stock-day. Appendix~\ref{app:generation} provides detailed failure counts and scoring flow.

\begin{table}[t]
 \centering
 \footnotesize
 \setlength{\tabcolsep}{4pt}
 \renewcommand{\arraystretch}{0.96}
 \begin{tabular}{@{}lrrr@{}}
 \toprule
 Self-conditioning condition & Scored $N$ & Timing alpha (95\% CI) & IC (95\% CI) \\
 \midrule
 No directive, $w=5$ & 1,289 & $-59.9\,[-67.1,-53.0]$ & $-0.0105\,[-0.0233,0.0017]$ \\
 Conviction, $w=5$ & 1,011 & $-58.1\,[-67.6,-48.6]$ & $+0.0028\,[-0.0082,0.0128]$ \\
 Consistency, current only ($w=0$) & 1,194 & $-62.8\,[-69.7,-56.0]$ & $-0.0134\,[-0.0278,0.0009]$ \\
 Consistency, one earlier ($w=1$) & 1,054 & $-74.1\,[-81.5,-66.6]$ & $-0.0125\,[-0.0260,0.0011]$ \\
 Consistency, $w=5$ & 1,154 & $-66.7\,[-73.9,-59.7]$ & $-0.0087\,[-0.0212,0.0036]$ \\
 Consistency, $w=12$ & 1,184 & $-67.1\,[-75.0,-59.6]$ & $-0.0076\,[-0.0190,0.0039]$ \\
 Multimodal consistency & 1,169 & $-64.8\,[-73.6,-56.0]$ & $+0.0002\,[-0.0130,0.0134]$ \\
 Multimodal, no directive & 1,274 & $-55.1\,[-62.4,-48.0]$ & $-0.0137\,[-0.0261,-0.0021]$ \\
 Encoder-off replication & 1,049 & $-70.2\,[-77.9,-62.5]$ & $-0.0042\,[-0.0165,0.0081]$ \\
 Encoder-word injection & 1,019 & $-51.5\,[-58.8,-44.5]$ & $-0.0110\,[-0.0227,0.0011]$ \\
 \midrule
 \multicolumn{4}{@{}l}{\textit{Later-period persistence and position changes}} \\
 Condition & Scored $N$ & Later-period alpha & Switches per stock-day \\
 \midrule
 No directive, $w=5$ & 277 & $-65.3$ & $2.5$ \\
 Consistency, $w=0$ & 270 & $-68.3$ & $2.5$ \\
 Consistency, $w=1$ & 245 & $-85.6$ & $1.7$ \\
 Consistency, $w=5$ & 252 & $-82.9$ & $1.8$ \\
 Consistency, $w=12$ & 251 & $-79.3$ & $1.9$ \\
 \bottomrule
 \end{tabular}
 \caption{Narrative self-conditioning results. Every main-period condition attempts 1,500 stock-days across 150 dates. Scored $N$ excludes failed intervals and constant action trajectories. The upper panel reports conditional timing alpha in bps per stock-day and contemporaneous cross-sectional IC with 95\% date-clustered confidence intervals. The lower panel reports later-period point estimates and observed long/flat position changes.}
 \label{tab:memory-sweep}
\end{table}

The timing estimates vary by more than 20~bps across the sweep, while nine of ten IC intervals contain zero. The multimodal no-directive condition has IC $-0.0137$ with interval $[-0.0261,-0.0021]$, yet its timing estimate is weaker than the consistency condition with one earlier memory. Timing therefore captures when a stock is held along its path, while \eqref{eq:IC} captures contemporaneous selection across stocks.
The lower panel records persistence alongside conditional timing. Position changes count long/flat transitions within each scored stock-day. The condition with one earlier memory ($w=1$) has the fewest changes and the most negative conditional timing in both periods. Longer windows remain more persistent than the current-memory condition ($w=0$), while their timing estimates move toward it. These action traces associate access to one earlier self-authored memory with greater persistence.
For the condition-level comparison below, cross-sectional portfolio return is the mean subsequent return across stocks held long at each decision time. Each condition contributes one timing estimate, one portfolio-return estimate, and one IC estimate.

\begin{table}[t]
 \centering
 \small
 \begin{tabular}{@{}lrrrr@{}}
 \toprule
 & \multicolumn{2}{c}{Conventional} & \multicolumn{2}{c}{Self-conditioning} \\
 Metric pair & Pearson & Spearman & Pearson & Spearman \\
 \midrule
 Timing vs portfolio return & $+0.355$ & $+0.392$ & $-0.066$ & $+0.018$ \\
 Timing vs cross-sectional IC & $+0.684$ & $+0.399$ & $-0.075$ & $-0.212$ \\
 Portfolio return vs cross-sectional IC & $+0.859$ & $+0.881$ & $+0.920$ & $+0.867$ \\
 \bottomrule
 \end{tabular}
 \caption{Condition-level metric correlations. Conventional contains 12 standard conditions. Self-conditioning contains ten memory conditions. Each condition contributes one observation.}
 \label{tab:metric-coupling}
\end{table}

Table~\ref{tab:metric-coupling} sharpens the distinction between timing and IC. Pearson correlations of timing with portfolio return are $+0.355$ conventionally and $-0.066$ under memory, while correlations with IC are $+0.684$ and $-0.075$. Portfolio return remains strongly associated with IC at $+0.859$ and $+0.920$, with the same ordering under Spearman correlation. Thus memory changes path-wise timing along a separate empirical axis from contemporaneous selection. Appendix~\ref{app:latent-details} provides additional condition-level interpretation.
Literal injection of the encoder's highest-weight words scores $-51.5$~bps, compared with $-70.2$ for the encoder-off replication in Table~\ref{tab:memory-sweep}. This contrast separates lexical rendering from embedding-space steering. The latent channel supplies a secondary continuous-state comparison. Table~\ref{tab:latent-ablation} reports the complete ablation record. The first group shares one cross-sectionally centered training outcome, while the second group records separate transformed objectives and supports descriptive sensitivity checks.

\begin{table}[t]
 \centering
 \small
 \setlength{\tabcolsep}{5pt}
 \begin{tabular}{@{}lr@{}}
 \toprule
 Configuration & Timing score, bps (95\% CI) \\
 \midrule
 \multicolumn{2}{@{}l}{\textit{Shared centered-outcome group}} \\
 Soft tokens disabled & $-21.0\,[-23.8,-18.3]$ \\
 Two tokens, 92-word basis & $-33.7\,[-37.5,-30.0]$ \\
 Eight tokens, 92-word basis & $-32.3$ \\
 Sixteen tokens, 92-word basis & $-29.2$ \\
 Two tokens, 492-word basis & $-33.8\,[-37.5,-30.1]$ \\
 \midrule
 \multicolumn{2}{@{}l}{\textit{Separate objectives or evaluation frames}} \\
 Trainable encoder, 92-word basis & $-38.4\,[-43.6,-33.3]$ \\
 Frozen encoder, 54k-parameter adapter & $-39.7\,[-46.8,-32.4]$ \\
 Free embedding vectors & $-2.1\,[-6.5,+2.6]$ \\
 Combined outcome transformation & $-28.7\,[-32.3,-25.0]$ \\
 Two-pass memory-conditioned encoder & $-15.4\,[-24.6,-5.1]$ \\
 \bottomrule
 \end{tabular}
 \caption{Latent-state ablations. Comparisons within the upper group share a training outcome. Lower-group rows use separate transformed objectives or evaluation frames. Date-clustered 95\% confidence intervals appear where available.}
 \label{tab:latent-ablation}
\end{table}

Within the shared centered-outcome group, two tokens shift timing from $-21.0$ to $-33.7$~bps relative to the disabled-token prompt. Eight and sixteen tokens yield $-32.3$ and $-29.2$~bps, while a 492-word basis yields $-33.8$~bps. Across the separate frames, trainable and frozen encoders score $-38.4$ and $-39.7$~bps, free embeddings score $-2.1$~bps, and the combined transformation scores $-28.7$~bps. The memory-conditioned checkpoint is fixed before evaluation on 1,500 stock-days from 28 later dates. It scores $-15.4$~bps with 95\% CI $[-24.6,-5.1]$, exposure $0.397$, and $6.71$ position changes per stock-day.
Appendix~\ref{app:latent-details} provides detailed arm definitions and checkpoint provenance.

\paragraph{Trace-level example.}
Table~\ref{tab:trace-examples} makes the timing estimand concrete with one standard price-text stock-day and one consistency stock-day with an earlier memory visible ($w=1$). In the second trace, the agent stays flat for decisions 1 to 7, switches long at decision 8, and remains long through decision 23. Its preceding memory states, ``My bullish thesis remains fully intact. The decisive breakout above 108.71 is confirmed, and the recent pullback to 114.39 served only as healthy consolidation.'' The numbers are normalized price levels from the observed history.

\begin{table}[t]
 \centering
 \small
 \setlength{\tabcolsep}{5pt}
 \begin{tabular}{@{}lrrrrr@{}}
 \toprule
 Condition & Long intervals & Exposure & Long return & Flat return & Timing alpha \\
 \midrule
 Price text & $7/23$ & $0.304$ & $-576.7$ & $+575.9$ & $-576.4$ \\
 Consistency, one earlier memory & $16/23$ & $0.696$ & $+359.4$ & $+1{,}404.0$ & $-867.3$ \\
 \bottomrule
 \end{tabular}
\caption{Trajectory examples. Returns and timing alpha are in bps under the research-return labels. The rows illustrate the estimand on stock-days and support the evidence in Tables~\ref{tab:main-results} and~\ref{tab:memory-sweep}.}
 \label{tab:trace-examples}
\end{table}

The price-text trajectory is long for seven intervals whose returns sum to $-576.7$~bps. The memory-conditioned trajectory earns a positive return while long, yet flat intervals capture a much larger part of the stock-day rise. Exposure centering assigns the second trajectory $-867.3$~bps of timing alpha. Its language preserves the bullish interpretation while its actions contain one position change. These cases show how path-wise timing reveals entry and exit quality beyond the long-only outcome.
Appendix~\ref{app:trajectory-details} provides the detailed archived memory and action summary.

\subsection{Behavioral signal and scope}
\label{sec:counter-policy}

RetailAgent identifies a predictable response pattern that another participant could diagnose from the agent's fixed-path action trace. Market theory shows how prices aggregate information and order flow conveys information to other participants~\citep{grossman1980impossibility,kyle1985continuous}. Historical evidence links small-trader demand to predictable subsequent returns~\citep{hvidkjaer2008small} and active individual trading to underperformance~\citep{barber2000trading}. The complementary schedule exchanges long and flat actions and serves as a sign diagnostic, while Section~\ref{sec:overlap} measures its linear overlap with three price-based signals.

Our evidence concerns a controlled LLM policy over anonymized historical equity paths with fixed prices and research-return labels. Matched human decision traces and executable-return analysis remain separate empirical requirements. Future work can compare LLM and human decisions under a matched interface and extend RetailAgent to interactive markets that incorporate transaction costs, borrowing constraints, latency, market impact, capacity, strategic counterparties, and price feedback. Magnitude comparisons across different sampling frames remain descriptive.
Appendix~\ref{app:validity} provides additional validity conditions and reporting conventions.

\section{Conclusion}
\label{sec:conclusion}

In this paper, we developed RetailAgent, a controlled framework for interpreting sequential LLM trading behavior. Across configurations spanning input modality, decision horizon, account state, and model family, agents exhibit consistently negative exposure-matched within-stock timing. Trajectory shuffling substantially attenuates this effect, while self-conditioning increases policy persistence and shifts conditional timing among stock-days with both actions. These findings reveal stable directional structure in LLM action traces beyond average exposure and the tested price-based signals. Future work can examine whether this structure remains actionable under transaction costs, market impact, and strategic feedback.

\clearpage
\bibliography{references}
\bibliographystyle{plainnat}

\clearpage
\appendix
\raggedbottom
\setcounter{figure}{0}
\setcounter{table}{0}
\renewcommand{\thefigure}{A\arabic{figure}}
\renewcommand{\thetable}{A\arabic{table}}
\renewcommand{\theHfigure}{A.\arabic{figure}}
\renewcommand{\theHtable}{A.\arabic{table}}
\setlength{\parskip}{3pt}

\section*{Supplementary Material}

\noindent
These appendices provide the information needed to interpret and reproduce the main results. Appendix~\ref{app:protocol} records the experimental protocol and reproducibility information. Appendix~\ref{app:additional} presents visual summaries and supporting analyses. The main paper retains the principal numerical comparisons, while these records document how each estimate was produced.

\section{Expanded experimental protocol}
\label{app:protocol}

\subsection{Observation panels and sequential state}
\label{app:panels}

CSI-500 is a mid-cap Chinese equity index~\citep{csi500}. The source data represent each anonymized stock-day on a 239-minute intraday grid. Numerical conditions receive an indexed serialization that preserves relative price movements while concealing the absolute price level. The primary local panel contains 23 scored 10-minute intervals per stock-day. The 5-minute and 20-minute panels contain 47 and 11 intervals. Their construction leaves terminal remainders of 4, 9, and 19 minutes outside the respective scored grids. Each outcome is the subsequent non-overlapping research-return label for the selected interval.

Numerical histories are constructed from previously completed non-overlapping intervals. Fine-resolution charts and encoder features use the archived one-minute series reconstructed from nested forward-return labels. That reconstruction depends on later label rows, so the fine-chart and learned-encoder conditions characterize behavior under the archived reconstructed input. The numerical-text conditions provide the direct information-boundary evidence based on previously completed intervals.

The standard configurations use a one-pass decision policy based on the current observation and condition-specific account state. Stateful prompts progressively add current position and entry price, a recent trade ledger, or an account summary of realized and unrealized performance. These variables arise from earlier actions and therefore create a recursive policy state. The two-pass narrative conditions first generate a short memory addressed to the model's future self. The second pass receives the current observation, the new memory, and up to $w$ earlier memories before selecting long or flat. Thus, $w=0$ includes the newly written memory and zero earlier memories, while $w\geq1$ adds prior memory history.

\subsection{Prompt templates}
\label{app:prompts}

The principal local numerical condition uses the following user prompt before application of the checkpoint's chat template. Bracketed fields are filled from the current stock-day and decision interval.

\begin{quote}
\small\raggedright
\texttt{You are a retail investor watching one stock during the trading day.}\\
\texttt{You decide once every [horizon] minutes whether to be in the market or out.}\\[2pt]
\texttt{For each decision:}\\
\texttt{1 = be in the market (long)}\\
\texttt{0 = be out of the market (flat)}\\[2pt]
\texttt{Reply in exactly this format, with no other text and no markdown:}\\
\texttt{MEMO: <one or two sentences of reasoning>}\\
\texttt{ACTION: <0 or 1>}\\[2pt]
\texttt{The final line MUST be exactly ``ACTION: 0'' or ``ACTION: 1''.}\\[2pt]
\texttt{Price history ([horizon]-minute bars, oldest to newest, index starts at 100):}\\
\texttt{[up to 20 observed price values]}\\[2pt]
\texttt{This is bar [t] of [T]. Your decision:}
\end{quote}

Stateful conditions retain this task and response format. The position condition adds current long or flat status, entry price when long, current indexed price, unrealized return, holding duration, and an accumulated action-weighted return labeled as realized profit and loss in the archived prompt. The ledger condition additionally presents the eight most recent closed trades with entry, exit, return, and holding duration. The account condition also presents the five most recent self-authored memos. These fields are updated from earlier actions and indexed prices. The accumulated return includes the contribution of an open long position, while the separate unrealized field measures the change from its entry price.

The chart-only and joint conditions pass a structured multimodal user message to the checkpoint processor. The image block precedes the text block, and the processor serializes it schematically as follows:
\begin{quote}
\small\raggedright
\texttt{user.content[0]: type=image}\\
\texttt{user.content[1]: type=text, text=[task instructions and current observation]}\\
\verb+<|vision_start|>+ \texttt{[image placeholder tokens]} \verb+<|vision_end|>+
\end{quote}
The implementation supplies the image as multimodal data, and the checkpoint processor generates the exact vision tokens through its chat template. The chart-only text states, ``The chart above shows the price history ([horizon]-minute bars, left to right).'' The joint condition attaches the rolling one-minute chart and includes the same numerical list used by the text condition. Both passes of a multimodal narrative condition attach the current chart before the corresponding memory or decision text. All multimodal prompts retain the common long and flat definitions and response format.

Narrative conditions make two calls per interval. The first call appends, ``First, write a short note to your later self about the market situation and your position. Begin the note with `NOTE:'.'' The consistency condition further appends, ``When you write your note, keep it consistent with the view in your earlier notes unless the evidence has clearly changed.'' The second call includes the newly generated note and appends, ``Now reply with a single character, 0 or 1. Your decision:'' The memory window $w$ controls only the number of earlier notes displayed in addition to the current note.

\subsection{Generation and parsing}
\label{app:generation}

The principal conditions use the locally cached \texttt{Qwen/Qwen3.5-9B} checkpoint~\citep{qwen35_9b} in bfloat16. Agents run through vLLM~\citep{kwon2023efficient} with the model's Hugging Face chat template, prefix caching, and thinking disabled. Numerical text and state conditions use temperature 0.7, top-$p$ 0.95, and a 160-token output limit. Chart and joint text-chart conditions use the same sampling parameters with a 320-token limit. Narrative memory generation uses temperature 0.7, top-$p$ 0.95, and a 220-token limit. Its second-pass binary decision is greedy with a 24-token limit. Hosted Haiku and Granite calls use temperature 0.7 and a 350-token limit. Local numerical generation uses two-way tensor parallelism. Stateful, chart, joint, and narrative generation use four-way tensor parallelism. Generation is batched across stock-days at each decision time, while decisions within each stock-day remain sequential.

Parsing follows the output format used by each generation path. Local numerical and state conditions select the final explicit \texttt{ACTION: 0/1} match and then apply directional-keyword fallback rules. Narrative decisions select the final standalone binary digit and use the same directional fallback. Hosted conditions select the first explicit action match. Local numerical and state failures map to flat. Chart and joint conditions retain a stock-day when every interval parses. Narrative scoring removes failed intervals before exposure centering. Hosted Haiku and Granite failures retain the incoming position and increment their recorded counters. The generation code contains the complete templates and parsing rules, while saved trajectories preserve actions and selected memory excerpts.

Table~\ref{tab:parse-failures} traces the narrative sample from generation to scoring. Each condition begins with 34,500 attempted decisions from 1,500 stock-days and 23 intervals. The archived scorer removes failed intervals, requires at least five valid decisions, and retains stock-days that contain both long and flat actions.

\begin{table}[t]
 \centering
 \footnotesize
 \setlength{\tabcolsep}{3.5pt}
 \begin{tabular}{@{}lrrrr@{}}
 \toprule
 Narrative condition & Failed intervals & Affected cells & Constant paths & Scored $N$ \\
 \midrule
 No directive, $w=5$ & 1,120 & 641 & 211 & 1,289 \\
 Conviction, $w=5$ & 1,690 & 584 & 489 & 1,011 \\
 Consistency, $w=0$ & 1,581 & 925 & 306 & 1,194 \\
 Consistency, $w=1$ & 1,545 & 615 & 446 & 1,054 \\
 Consistency, $w=5$ & 505 & 344 & 346 & 1,154 \\
 Consistency, $w=12$ & 266 & 207 & 316 & 1,184 \\
 Multimodal consistency & 425 & 186 & 331 & 1,169 \\
 Multimodal, no directive & 233 & 147 & 226 & 1,274 \\
 Encoder-off replication & 1,657 & 608 & 451 & 1,049 \\
 Encoder-word injection & 1,407 & 609 & 481 & 1,019 \\
 \bottomrule
 \end{tabular}
 \caption{Narrative scoring provenance. Affected cells contain at least one failed interval. Constant paths remain all long or all flat after failed intervals are removed. The archived conditional timing estimates use the scored stock-days in the final column.}
 \label{tab:parse-failures}
\end{table}

\paragraph{Prompt-horizon provenance.}
The interval column in the results tables records the panel and scoring cadence. The archived local state generator maps the 47-bar panel to 5-minute wording and assigns 10-minute wording to the 23-bar and 11-bar panels. Its 20-minute state conditions therefore display 10-minute wording. The visual generator computes the displayed horizon as $239/T$, which produces 21-minute wording on the 11-bar panel. The hosted prompt uses fixed 10-minute wording, including the archived 5-minute Haiku condition. These wording differences qualify comparisons across horizon rows while leaving each row's scoring interval unchanged.

\subsection{Encoder optimization and chronological split}
\label{app:encoder-optimization}

The price encoder is a two-layer GRU~\citep{cho2014learning} with 256 hidden units, a 256-dimensional output, dropout 0.1, and a 120-minute lookback. The archived implementation stacks one-minute return, 10-minute volatility, 30-minute volatility, a second adjacent-return difference stored under the \texttt{mom5} name, 20-minute momentum, drawdown, and time of day. VICReg pretraining~\citep{bardes2021vicreg} uses invariance and variance weights of 25 and covariance weight 1, together with 10\% magnitude jitter and 10\% channel dropout. Pretraining runs for eight epochs with batch size 512 on 60 stocks selected with NumPy seed 0. AdamW uses learning rate $3\times10^{-4}$ and weight decay $10^{-4}$.

The principal adapter uses two soft tokens, a rank-64 bottleneck, and a 92-token lexical basis. The phase-two runs use AdamW~\citep{loshchilov2018decoupled} with default learning rates of $10^{-4}$ for the adapter and $10^{-5}$ for the encoder. The frozen-encoder run uses adapter learning rate $3\times10^{-4}$. Phase two initializes a fixed decision bias and applies an exposure-collapse penalty outside the interval $[0.10,0.60]$. The memory-conditioned phase-three run uses learning rates of $3\times10^{-4}$ for the adapter and $10^{-5}$ for the encoder, together with an online scalar bias targeting exposure 0.40. The principal phase-two ablations use 100 optimizer steps and 480 stock-days per step. Training dates occupy the first three chronological fifths of the relevant panel. The saved memory-conditioned checkpoint is evaluated on 1,500 stock-days from 28 later dates. The narrative sweep attempts 1,500 stock-days over 150 dates.

Panel construction and statistical evaluation use recorded seeds. The primary panel samples 150 dates and up to 200 stocks per date with seed 4242. The 5-minute panel samples 100 dates with seed 20260811 and 100 stocks per date with date-specific seed $20260810+d$. The cross-sectional panel uses seed 31337. Narrative conditions select 1,500 stock-days with seed 0. Learned-condition replication treats the archived checkpoints as canonical because the training scripts use framework-level randomness and omit a global training seed.

\subsection{Inference and metric provenance}
\label{app:inference}

Timing alpha first aggregates over intervals within each stock-day and then averages across the stock-days admitted by the corresponding scorer. Standard and narrative scoring retain trajectories with both long and flat actions. Equation~\eqref{eq:alpha} assigns zero timing to an all-long or all-flat trajectory, so the reported means characterize the switching subset. Table~\ref{tab:parse-failures} distinguishes the 1,500 attempted narrative stock-days from the scored counts. Confidence intervals use a date-clustered percentile bootstrap that resamples dates with seed 0. Standard-arm validation and signal-overlap analyses use 3,000 draws. Narrative analyses use 2,000 draws. Hosted-model summaries use 5,000 draws. Cross-sectional IC averages over decision times with cross-sectional position variation. Constant-position cross-sections are undefined and omitted.

For the overlap analysis, the inverted position is $q=1-p$. For each comparator $s$, the analysis projects the within-stock centered $q$ onto centered $s$ and evaluates the residual schedule. This procedure measures linear overlap with the tested comparator. The GRU row uses its available 2,519-stock-day evaluation frame, while GBDT-38 and one-period reversal use 13,717 stock-days. Comparisons across these frames are descriptive.

The later-period narrative panel uses dates 296 through 354. The consistency directive was selected using the broader panel that includes these dates. The later-period rows therefore provide a chronological extension of the observed pattern rather than a directive-selection holdout.

\subsection{Reproducibility record}
\label{app:reproducibility}

The accompanying artifact contains panel-construction and scoring code, the prompt implementations summarized in Appendix~\ref{app:prompts}, aggregate trajectories, per-cell hosted-model records, encoder checkpoints, failure counters, and machine-readable evaluation outputs. Saved positions directly reproduce the reported scores. Generation code and aligned panel values reconstruct the prompts. Narrative trajectory files preserve parsed actions and one truncated midpoint memory excerpt per stock-day. Checkpoint arguments and filenames preserve most model identifiers, while aligned panels provide the subsequently revealed labels. Appendix~\ref{app:model-provenance} records the remaining coarse-image checkpoint ambiguity.

Scoring requires NumPy, Polars, and scikit-learn on CPU. Regenerating local model actions additionally requires PyTorch, Transformers, vLLM, and the relevant model weights. The compute cluster uses NVIDIA H100 GPUs with 80~GB HBM3 memory. Local numerical inference uses two GPUs, and stateful, visual, multimodal, and narrative inference uses four GPUs. The 1,500-stock-day headline memory arm takes approximately one hour on four H100 GPUs. Encoder and adapter training uses eight H100 GPUs. Eleven completed archived training traces total 39,885 seconds of wall time, corresponding to 88.6 H100 GPU-hours. This quantity is a documented lower bound. A full project total additionally requires scheduler accounting for generation, pilot runs, and interrupted jobs.

The hosted-model cache records 218,639 calls, 98,041,696 prompt tokens, and 15,322,815 completion tokens across the archived Haiku and Granite sweeps. Gateway billing records determine the corresponding monetary cost. The artifact records panel and evaluation seeds as described above. Exact software versions, immutable model revision hashes, CPU allocation, system memory, and full-project compute accounting remain items for the final release record.

\section{Additional experimental and analysis details}
\label{app:additional}
\setcounter{figure}{0}
\renewcommand{\thefigure}{B\arabic{figure}}
\renewcommand{\theHfigure}{B.\arabic{figure}}

\begin{figure}[t]
 \centering
 \includegraphics[width=0.88\linewidth]{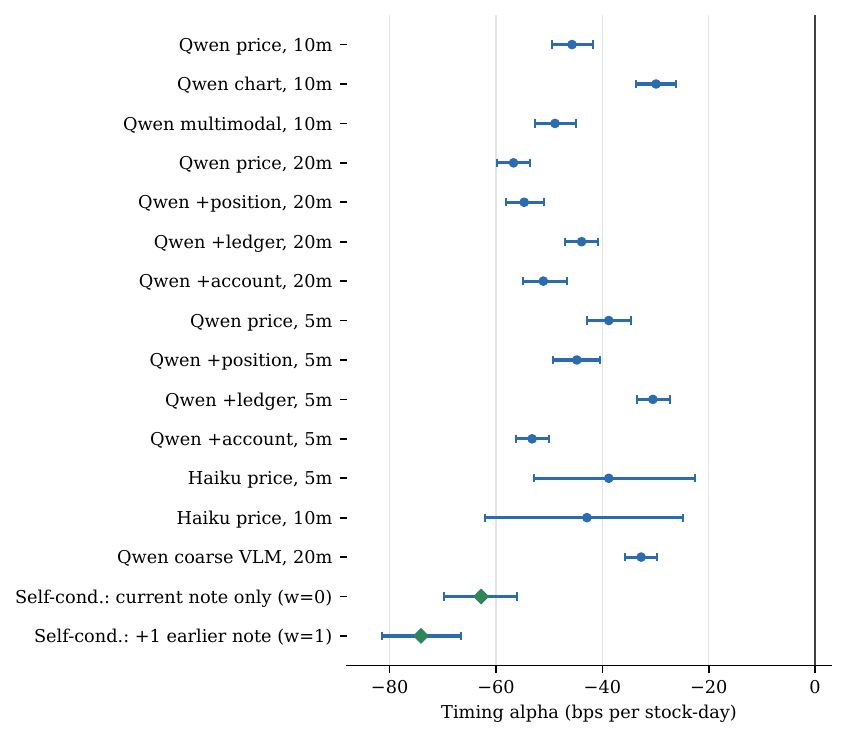}
 \caption{Standard and representative self-conditioning timing estimates for the locally cached \texttt{Qwen/Qwen3.5-9B} checkpoint and the archived Qwen3-VL coarse-image condition~\citep{qwen35_9b,bai2025qwen3vl}, with the Claude Haiku 4.5 sign check~\citep{anthropic2025haiku}. Points report within-stock timing alpha with 95\% date-clustered confidence intervals. Table~\ref{tab:main-results} reports sample sizes and exact values.}
 \label{fig:app-standard}
\end{figure}

\begin{figure}[t]
 \centering
 \includegraphics[width=0.88\linewidth]{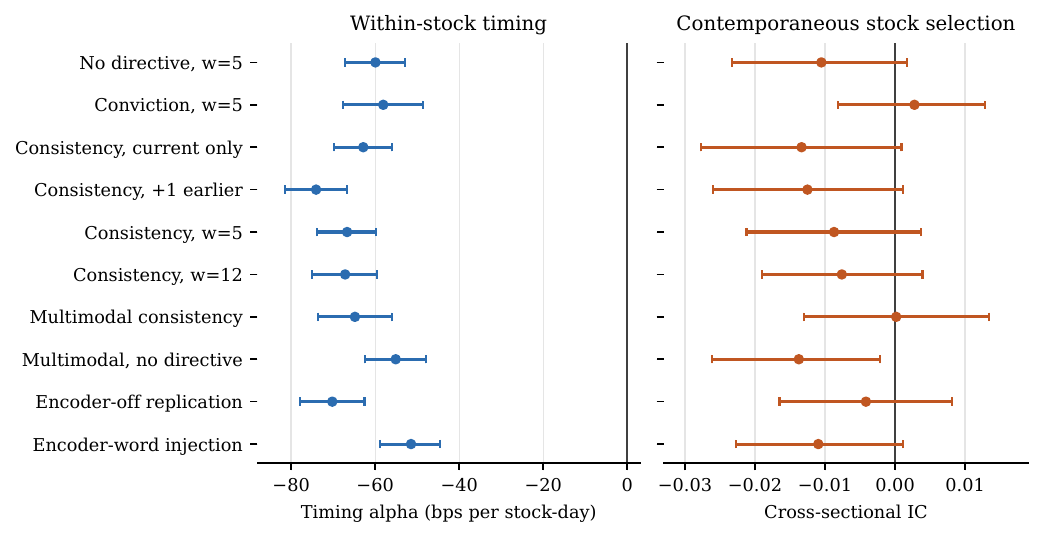}
 \caption{Within-stock timing alpha and contemporaneous cross-sectional IC for ten Qwen3.5-9B narrative self-conditioning conditions~\citep{qwen35_9b}. Each condition attempts 1,500 stock-days. Scored counts appear in Table~\ref{tab:memory-sweep}. Both panels report 95\% date-clustered confidence intervals.}
 \label{fig:app-memory}
\end{figure}

Figures~\ref{fig:app-standard} and~\ref{fig:app-memory} provide visual counterparts to Tables~\ref{tab:main-results} and~\ref{tab:memory-sweep}. The tables remain the numerical record. The figures make sign recurrence, uncertainty, and the separation between timing and cross-sectional selection easier to inspect.

\subsection{Latent-state and metric analyses}
\label{app:latent-details}

Table~\ref{tab:latent-ablation} separates configurations by objective comparability. The disabled-token row, the two-token principal arm, the eight-token and sixteen-token variants, and the 492-word basis variant share one cross-sectionally centered outcome. The trainable-encoder, frozen-encoder, free-vector, combined-transformation, and memo-conditioned rows use separate transformed objectives or evaluation frames. Their values document the archived procedures within those groups.

The disabled-token row receives a numerical price list. The token-enabled phase-two rows receive price state through soft tokens as their sole price representation. The archived evaluations therefore compare input channels under a shared outcome rather than isolating token presence while holding the prompt representation fixed. The principal constrained arm uses two tokens formed on the 92-word lexical basis. The eight-token and sixteen-token variants change token bandwidth. The 492-word variant changes the lexical basis size. The frozen-encoder condition trains only the 54k-parameter adapter. The free-vector condition removes the lexical constraint. The combined-transformation condition uses its separately recorded outcome transformation. The code contains a matched price-text-plus-token option, and the archive currently contains the channel comparison reported in Table~\ref{tab:latent-ablation}.

The memo-conditioned checkpoint is trained inside the two-pass narrative procedure and evaluated on 1,500 stock-days from 28 later dates. Its prompt, sample, and training procedure define a separate frame from the narrative-only conditions. The stored centered score is $-15.0$ with interval $[-20.3,-9.1]$. Rescoring the same positions on raw returns produces the reported $-15.4$ value with interval $[-24.6,-5.1]$. The launch configuration specifies 12 steps, its progress trace records steps 1 through 8, and the checkpoint omits a step field. We treat the saved checkpoint as the canonical evaluated state and leave its stopping iteration unspecified.

\paragraph{Cross-condition metric interpretation.}
\label{app:metric-details}

Each observation in Table~\ref{tab:metric-coupling} is one experimental condition with complete trajectories. Portfolio return is the mean subsequent return across stocks held long at each decision time. Timing and IC have Pearson correlation 0.684 across 12 conventional conditions and $-0.075$ across the ten narrative conditions. Portfolio return and IC have corresponding correlations of 0.859 and 0.920. These condition-level summaries describe metric coupling rather than stock-day sampling uncertainty.

\subsection{Archived trajectory and cross-model records}
\label{app:trajectory-details}

Table~\ref{tab:trace-examples} contains ex-post illustrations of \eqref{eq:alpha}. In the price-text example, seven long intervals sum to $-576.7$ bps, sixteen flat intervals sum to $+575.9$ bps, and timing alpha equals $-576.4$ bps. In the $w=1$ consistency example, the agent switches long at decision 8 and remains long through decision 23 after preserving its bullish interpretation in memory. The trajectory has exposure 0.696, one position change, long-interval return of $+359.4$ bps, flat-interval return of $+1{,}404.0$ bps, and timing alpha of $-867.3$ bps. The aggregate tables provide the principal evidence.

\paragraph{Cross-model record provenance.}
\label{app:model-provenance}

Table~\ref{tab:main-results} contains the complete cross-model sign checks. Claude Haiku 4.5~\citep{anthropic2025haiku} produces negative estimates at both tested horizons, with 827 stock-days at 5 minutes and 165 stock-days at 10 minutes. Granite 4.0~\citep{granite2025} produces negative point estimates across price-only, position, ledger, and account prompts on 600 stock-days per condition.

The Granite point estimates were recovered from per-cell records after a cache migration. The cache preserves actions, failure counters, and the 600-stock-day condition summaries, while the archived condition summaries provide point estimates only. Hosted failures retain the incoming position. At least one API failure occurs in 33 price-only stock-days, 76 position stock-days, 8 ledger stock-days, and zero account stock-days. The point estimates in Table~\ref{tab:main-results} reproduce the archived 600-stock-day summaries with this recorded fallback behavior. Accordingly, the Granite evidence serves as a directional sign check, while the Qwen trajectories provide the principal evidence base.

The coarse-image trajectory archive is labeled Qwen3-VL, while the included visual launcher defaults to \texttt{Qwen/Qwen3.5-9B} and the saved trajectory lacks an embedded checkpoint identifier. The exact checkpoint for this row remains an author-verification item. Its reported timing value describes the archived trajectory regardless of the final model label.

\subsection{Validity conditions and reporting conventions}
\label{app:validity}

\textbf{Empirical scope.} RetailAgent characterizes a controlled LLM policy over anonymized historical paths. Its outcomes are research-return labels. Economic evaluation would incorporate executable prices, borrowing rules, latency, and capacity. Transaction costs and permanent and temporary market impact are central components of optimal-execution analysis~\citep{almgren2001optimal}. A market period reserved from design and selection would provide a selection-independent evaluation, while specialized procedures can quantify backtest-overfitting risk~\citep{bailey2017probability}. An interactive-market extension would model endogenous order flow and price feedback~\citep{kyle1985continuous}. Multi-agent LLM studies provide complementary frameworks for strategic interaction~\citep{piatti2024cooperate,sreedhar2024simulating}.

\textbf{Comparison scope.} Several conditions use different sampling frames, prompt wording, and evaluation procedures. We interpret their repeated signs as configuration-level evidence and compare magnitudes when prompt, sample, reward, and inference procedure coincide. The latent-state table identifies rows that share an outcome and rows that use separate objectives or evaluation frames.

\textbf{Implementation choices.} Phase-two learned runs use a fixed initial decision bias and an exposure-collapse penalty. The memory-conditioned phase-three run updates its bias online toward exposure 0.40. Reported learned runs estimate their outcome transformations on training dates and evaluate saved checkpoints on designated later observations. Hard binary actions define evaluation, while the training path uses the straight-through gradient estimator~\citep{bengio2013estimating}. The exposure controls keep learned policies away from all-long and all-flat boundaries.

\textbf{Metric conventions.} Timing alpha is a raw within-stock covariance in basis points per stock-day on its original scale. The archived standard and narrative scorers condition this average on trajectories containing both actions. The complementary schedule exchanges every long and flat action and serves as a sign diagnostic. Signal projections measure linear overlap with three tested comparators. Cross-sectional IC averages over valid decision times with cross-sectional position variation.

\end{document}